\documentclass[10pt,twocolumn,letterpaper]{article}
\usepackage[toc,page]{appendix}

\usepackage{cvpr}
\usepackage{times}
\usepackage{epsfig}
\usepackage{graphicx}
\usepackage{amsmath}
\usepackage{amssymb}
\usepackage{algorithm}
\usepackage{algpseudocode}
\usepackage{multirow}
\usepackage{booktabs} 
\usepackage{caption}
\usepackage{subcaption}
\usepackage{xcolor}

\usepackage[pagebackref=true,breaklinks=true,letterpaper=true,colorlinks=true,bookmarks=false]{hyperref}
\hypersetup{
    urlcolor=magenta,
}
\cvprfinalcopy 

\def\cvprPaperID{6959} 

\ifcvprfinal\fi
\begin{document}

\title{Towards Large yet Imperceptible Adversarial Image Perturbations with Perceptual Color Distance}

\author{Zhengyu Zhao, Zhuoran Liu, Martha Larson\\
Radboud University, Nijmegen, Netherlands\\
{\tt\small \{z.zhao, z.liu, m.larson\}@cs.ru.nl}
}

\maketitle

\begin{abstract}
The success of image perturbations that are designed to fool image classifier is assessed in terms of both adversarial effect and visual imperceptibility.
The conventional assumption on imperceptibility is that perturbations should strive for tight $L_p$-norm bounds in RGB space.
In this work, we drop this assumption by pursuing an approach that exploits human color perception, and more specifically, minimizing perturbation size with respect to perceptual color distance.
Our first approach, Perceptual Color distance C\&W (PerC-C\&W), extends the widely-used C\&W approach and produces larger RGB perturbations.
PerC-C\&W is able to maintain adversarial strength, while contributing to imperceptibility.
Our second approach, Perceptual Color distance Alternating Loss (PerC-AL), achieves the same outcome, but does so more efficiently by alternating between the classification loss and perceptual color difference when updating perturbations.
Experimental evaluation shows PerC approaches outperform conventional $L_p$ approaches in terms of robustness and transferability, and also demonstrates that the PerC distance can provide added value on top of existing structure-based methods to creating image perturbations.

\end{abstract}

\begin{figure}[t]
\begin{center}
  \includegraphics[width=\columnwidth]{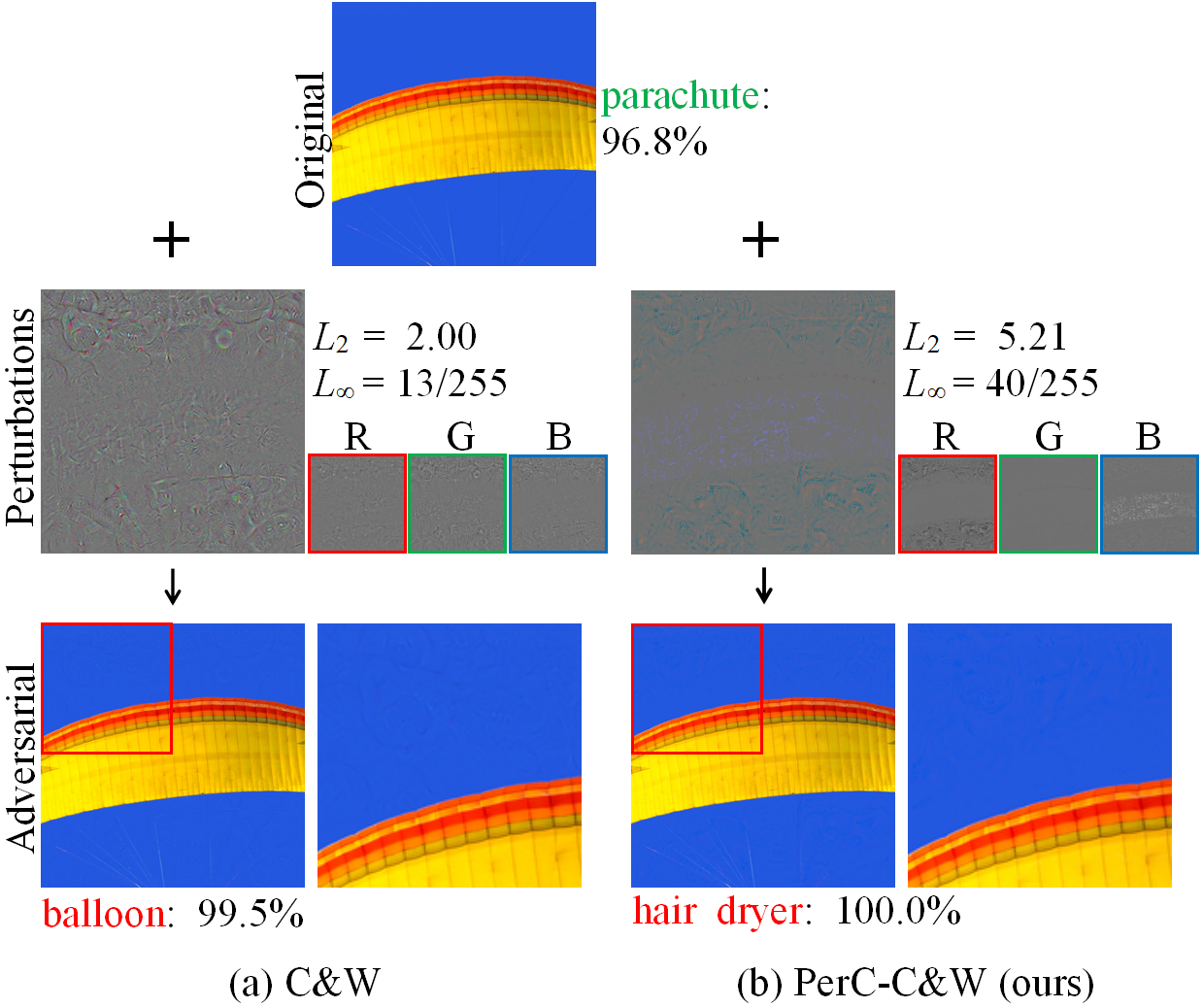}
\end{center}
   \caption{Comparison of (a) C\&W~\cite{carlini2017towards} with (b) our PerC-C\&W. Perceptual color (PerC) distance allows larger RGB perturbations (cf. $L_{2}$ and $L_{\infty}$ norm in middle row), while also contributing to imperceptibility (bottom row). (Setting: untargeted with $\kappa=40$; classifier Inception v3.)}
\label{fig:example}
\end{figure}

\section{Introduction}
\label{intro}
Research on creating adversarial examples for deep visual classifiers has focused on perturbations that cause misclassification while being \textit{imperceptible} to the human eye~\cite{carlini2017towards,papernot2016limitations,szegedy2013intriguing}.
Larger image perturbations are known to improve adversarial strength (i.e., the ability to fool a classifier), but are also associated with visually noticeable changes in the image. 
A commonly agreed-upon assumption is that tight $L_{p}$-norm constraints on the size of adversarial perturbations in RGB space are a good guarantee of imperceptibility.
Evaluation of adversarial examples has conventionally followed this assumption, considering perturbations with smaller $L_{p}$ norms to be better (e.g., $L_{\infty}$~\cite{carlini2017towards,goodfellow2014explaining,kurakin2016adversarial}, $L_{2}$~\cite{carlini2017towards,moosavi2016deepfool,szegedy2013intriguing} and $L_{0}$~\cite{carlini2017towards,papernot2016limitations}).
Keeping with this assumption, defense approaches are designed to be effective against adversarial perturbations under a specific $L_{p}$ bound~\cite{cohen2019certified,madry2017towards,tramer2017ensemble,wong2018provable}.
Our research is motivated by the importance of questioning the necessity of small RGB perturbations for imperceptibility.

In this work, we propose to create adversarial examples by perturbing images with respect to perceptual color (PerC) distance.
Using PerC distance makes it possible to move away from the assumption that it is necessary to tightly constrain the $L_{p}$ norm of the perturbations in RGB space. 
Fig.~\ref{fig:example} illustrates the difference between C\&W~\cite{carlini2017towards}, a well-known approach that perturbs with respect to an $L_{p}$ norm in RGB space, and our own extension, PerC-C\&W, which perturbs with respect to a perceptual color distance.
PerC perturbations are less perceptible, especially in smooth regions of saturated color (cf. Fig.~\ref{fig:example} in bottom row).  
Also, they are distributed strategically over the RGB color channels (cf. downsized perturbation images in the middle row). 
PerC distance effectively allows us to hide large perturbations in RGB space, in a way not readily noticeable to the human eye.
Our PerC-based approaches can increase the $L_{p}$ norm substantially (cf. Fig.~\ref{fig:example}, $L_{2}$ and $L_{\infty}$ in middle row), leading to a strong adversarial effect that maintains imperceptibility.

\begin{figure}[t]
\begin{center}
  \includegraphics[width=\columnwidth]{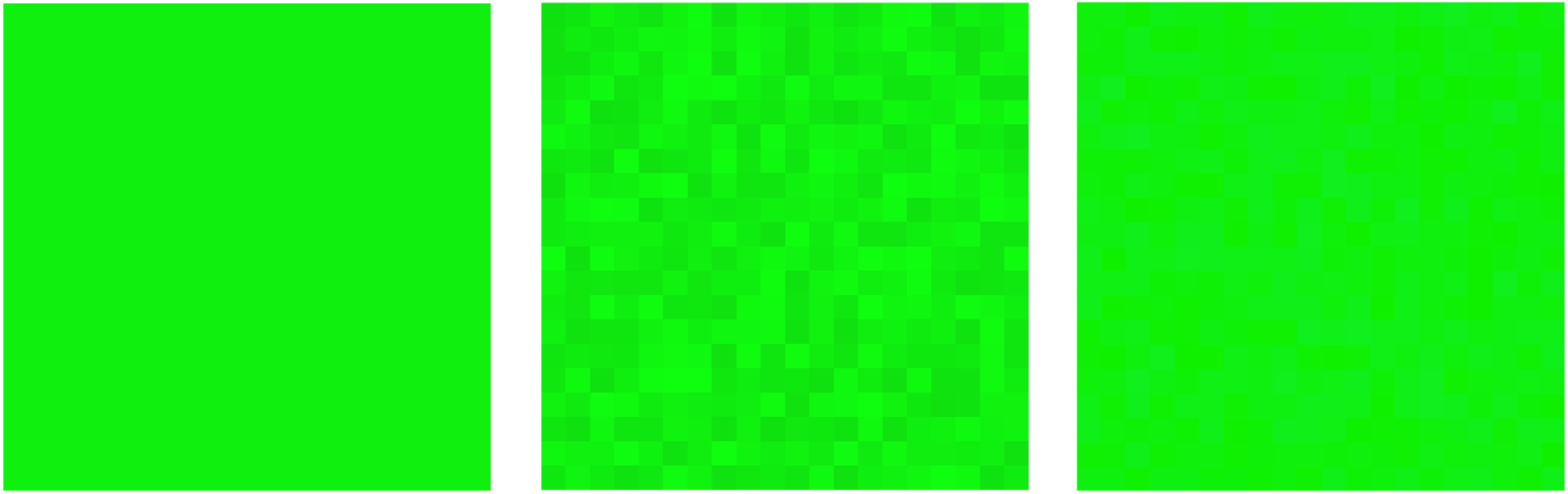}
\end{center}
   \caption{Left: Original image (a $20\times20$ 8-bit RGB image patch with color (15,240,15)). Middle: Image perturbed by adding noise in the G channel, sampled from a uniform distribution in the range [-15,15]. Right: Image perturbed by adding the identical noise, but in the B channel. The B-channel perturbations are imperceptible (best viewed on screen).}
\label{fig:idea}
\end{figure}

Fig.~\ref{fig:idea} motivates the use of perceptual color distance for creating adversarial images.
Here, we have taken a solid color image (left) and added the same perturbations to the green channel (middle) and to the blue channel (right). 
Although both RGB channels were perturbed identically, the perturbations are only visible in the green channel.
The reason is that color as it is perceived by the human eye does not change uniformly over distance in RGB space.
Relatively small perturbations in RGB space may correspond to large difference in perceptual color space.
Conversely, relatively large changes in RGB space may remain unnoticeable if they lead to small perceived color difference.

Our work is in line with a growing awareness in the literature on adversarial examples that the difference between two images as measured by an $L_{p}$ norm 
in RGB space 
is actually quite poorly aligned with human perception~\cite{sharif2018suitability}.
Building on this observation, researchers have attempted to address imperceptibility by exploiting similarity defined with respect to semantics~\cite{engstrom2017rotation,eykholt2017robust,hosseini2018semantic,joshi2019semantic,sharif2019adversarial} or structural information~\cite{croce2019sparse,gragnaniello2019perceptual,luo2018towards,wong2019wasserstein,zhang2020smooth} in the image.
However, little work on adversarial examples has questioned the wisdom of optimizing perturbations with respect to distance in RGB space.
The exceptions are a handful of approaches that have proposed allowing only luminance change when perturbing pixels~\cite{croce2019sparse,gragnaniello2019perceptual}.
The approach that is closest to our own is~\cite{athalye2018synthesizing}, which perturbs in CIELAB color space, but carries out no investigation of the potential and limitations of the idea.
Our work is distinct from this initial effort because we use a more accurate polar form (known as CIELCH) of the CIELAB color space, and more importantly, use an actual perceptual color distance.
The distance is CIEDE2000~\cite{ciede2000,luo2001development}, and will be discussed in detail in Section~\ref{sec:colordiff}.
To our knowledge, ours is the first work that proposes optimizing adversarial image perturbations directly with respect to a perceptual color distance.

In order to fully appreciate our proposal, it is necessary to understand two key aspects.
First, we do not claim that PerC approaches will always yield dramatically less perceptible perturbations than conventional RGB approaches.
For cases in which the perturbations are small, the difference may not be so great.
However, we find that there are two cases in which PerC approaches are particularly important.
First, our experimental results (see Section~\ref{sec:high}) show that as we attempt to create adversarial images that are misclassified with high confidence (i.e., high-confidence adversarial examples), it becomes important to perturb with respect to perceptual color distance.
Second, we demonstrate that the effect of PerC approaches is additive and can be used in combination with existing structural approaches to improve imperceptibility.

The contributions of this paper are as follows:
\vspace{-\topsep}
\begin{itemize}
  \setlength{\parskip}{0pt}
  \setlength{\itemsep}{0pt}
\item An in-depth study of the use of perceptual color (PerC) distance to hide large RGB perturbations in images.
\item PerC-C\&W: a method for creating adversarial images that introduces perceptual color distance into the joint optimization of C\&W.
\item PerC-AL: an efficient method that optimizes alternating loss (AL) functions, switching between classification loss and perceptual color difference.
\item Experimental validation demonstrating that PerC perturbations in high-confidence settings yield more robust and transferable adversarial examples, without sacrificing imperceptibility.
\item Experimental results showing that PerC perturbations can be used in combination with structural information for further improvement of imperceptibility.
\end{itemize}
\vspace{-\topsep}

The code, which also includes a differentiable solution compatible with PyTorch's autograd to efficiently implement perceptual color distance (CIEDE2000), is available at~\url{https://github.com/ZhengyuZhao/PerC-Adversarial}.

\section{Background on Perceptual Color Distance}
\label{sec:colordiff}

Conventionally, computer vision research has intensively explored color and human perception, but has paid surprisingly little attention to distance in perceptual color spaces.
Here, we mention some key points about color in computer vision history.
Early on, research focused on intensity-based descriptors, which then evolved to also capture color information. 
Unsurprisingly, color boosted the performance of object and scene recognition~\cite{khan2012color,van2009evaluating} and semantic segmentation~\cite{cheng2001color}.
Researchers extracted descriptors from opponent color spaces, most notably HSV and CIELAB, which separate luminance and chrominance.
Most recently, color is attracting more attention in the area of image synthesis. 
Notable examples, such as style transfer~\cite{gatys2016preserving} and cross-domain image generation~\cite{taigman2016unsupervised}, find that color plays an important role in preserving the look of an image.
In general, we observe that until now the focus has been on the color space itself, and not on color distance, which we explore here.

The perceptual color distance that we use is CIEDE2000, which is the latest $\Delta E$ standard formula developed by the CIE (International Commission on Illumination).
CIEDE2000 refined the definition of previous editions by adding five corrections, and has been experimentally demonstrated to better align with human visual perception~\cite{ciede2000,luo2001development}.
Specifically, the pixel-wise perceptual color distance can be calculated as:
\begin{equation}
\label{deltaE}
\begin{gathered}
\Delta E_{00}=\sqrt{(\frac{\Delta L'}{k_LS_L})^2+(\frac{\Delta C'}{k_CS_C})^2+(\frac{\Delta H'}{k_HS_H})^2+\Delta R},\\
\Delta R=R_T(\frac{\Delta C'}{k_CS_C})(\frac{\Delta H'}{k_HS_H}),
\end{gathered}
\end{equation}
where $\Delta L'$, $\Delta C'$, $\Delta H'$ denotes the distance between pixel values of the three channels, L (lightness), C (chroma) and H (hue) in the CIELCH space, and $\Delta R$ is an interactive term between chroma and hue differences~\cite{luo2001development}.
The weighting functions $S_L$, $S_C$, $S_H$ and $R_T$ are determined based on large-scale human studies and act as compensations to better simulate human color perception.
The $k_L$, $k_C$ and $k_H$ are usually unity for the application of graphic arts.
Detailed definitions of all the parameters and relevant explanations can be found in~\cite{luo2001development}.
We note that it is also possible to use an $L_{p}$ norm to measure distance in CIELAB space.
However, this distance is not as close to human perceptual distance as CIEDE2000 is.

We point out that a limited amount of previous research has also adopted CIEDE2000. However, the goal has been to evaluate the color similarity of image pairs.
Examples of such research include work on image quality assessment~\cite{yang2012color} and image super-resolution~\cite{liu2010colorization}.
In contrast, in our work we use CIEDE2000 directly for optimization with back propagation and not only for evaluation.

\section{Related work}
In this section, we cover the existing literature, which focuses on creating $L_p$ norm-bounded adversarial examples, and we also mention recent approaches that attempt to move beyond $L_p$ norms.
We preface our discussion with a short definition of an `adversary', i.e., an approach that generates an adversarial image example.
Given a classifier $f(\boldsymbol{x}):\boldsymbol{x} \to y$ that predicts a label $y$ for an image $\boldsymbol{x}$, the adversary attempts to induce a misclassification by modifying the original $\boldsymbol{x}$ to create a new $\boldsymbol{x}'$.
In the untargeted setting, the adversary is successful if the image is classified into an arbitrary class other than $y$, i.e., meets the condition $f(\boldsymbol{x}')\neq{y}$.
In the targeted setting, the adversary must ensure that the image is classified into a class with a pre-defined label $t$, i.e., meets the condition $f(\boldsymbol{x}')=t$.
The untargeted case is generally recognized to be less challenging than the targeted case~\cite{carlini2017towards}.

\subsection{$L_{p}$ norm-bounded Adversarial Examples}
\label{norm}
Typically, adversaries ~\cite{carlini2017towards,goodfellow2014explaining,kurakin2016adversarial,moosavi2016deepfool,papernot2016limitations,rony2019decoupling,szegedy2013intriguing} create an adversarial image, $x'$, by adding a perturbation vector
$\boldsymbol{\delta}\in\mathbb{R}^n$ that is constrained by an $L_{p}$ norm to the original image, $\boldsymbol{x}$.
The first $L_{p}$ norm-bounded approach~\cite{szegedy2013intriguing} optimized an objective combining the classification loss and the $L_{2}$ norm of the perturbations, balanced by a constant $\lambda$.
Formally, the solution is expressed as:
\begin{equation}
\label{opt}
\underset{\boldsymbol{\delta}}{\mathrm{minimize}}~\lambda{\|\boldsymbol{\delta}\|}_2-J(\boldsymbol{x}',y),~ \textrm{s.t.}{~\boldsymbol{x}'}\in[0,1]^n, 
\end{equation}
where $J(\boldsymbol{x}',y)$ is the cross-entropy loss w.r.t.~$\boldsymbol{x}'$. The authors of~\cite{szegedy2013intriguing} solved the problem by using box-constrained L-BFGS (Limited memory Broyden-Fletcher-Goldfarb-Shanno) method~\cite{liu1989limited}.

The C\&W method~\cite{carlini2017towards} improves on~\cite{szegedy2013intriguing} by introducing a new variable using the tanh function to eliminate the box constraint. 
Additionally, it introduces a more sophisticated objective function that optimizes differences between the logits, $Z$, which are output before the softmax layer.
This can be formulated as:
\begin{equation}
\label{cw}
\begin{aligned}
&\underset{\boldsymbol{w}}{\mathrm{minimize}}
~~{{\|\boldsymbol{x}'-\boldsymbol{x}\|}_2^2}+\lambda f(\boldsymbol{x}'),\\
&\textrm{where}~~ f(\boldsymbol{x}')=\mathrm{max}({\mathrm{max}}\{Z(\boldsymbol{x}')_i:i\neq t\}-Z(\boldsymbol{x}')_t,-\kappa),\\
&\textrm{and}~~\boldsymbol{x}'=\frac{1}{2}(\tanh(
\mathrm{arctanh}(\boldsymbol{x})+\boldsymbol{w})+1),
\end{aligned}
\end{equation}
where $\boldsymbol{w}$ is the new variable and $Z(x')_i$ denotes the logit with respect to the $i$-th class.
In an untargeted setting, the definition of $f$ is modified to:
\begin{equation}
\label{eq:cw_untar}
f(\boldsymbol{x}')=\mathrm{max}(Z(\boldsymbol{x}')_y-{\mathrm{max}}\{Z(\boldsymbol{x}')_i:i\neq y\},-\kappa).
\end{equation}
The parameter $\kappa$ controls the confidence level of the misclassification.
The first approach that we propose, PerC-C\&W, is built on C\&W.
In our experiments, we will vary $\kappa$ in order to assess the ability of an adversary to create strong adversarial images, i.e., images that are misclassified with high confidence. 

Due to the need for line search in order to find the optimal constant, $\lambda$, such an optimization approach is inevitably time-consuming.
For this reason,~\cite{goodfellow2014explaining,kurakin2016adversarial,rony2019decoupling} propose a more efficient solution that does not impose a penalty during optimization.
Instead, respect of the norm constraint is ensured by projecting perturbations onto an $\epsilon$-sphere around the original image.
Specifically, the fast gradient sign method (FGSM)~\cite{goodfellow2014explaining} was first proposed to achieve adversarial effect with only one step, formulated as:
\begin{equation}
\label{FGSM}
\boldsymbol{x}'=\boldsymbol{x}+\epsilon \cdot\mathrm{sign}(\nabla_{\boldsymbol{x}}J(\boldsymbol{x},y)),
\end{equation}
where the perturbation size is implicitly constrained by specifying a small $\epsilon$. 

Subsequently, an extension of this method referred to as I-FGSM~\cite{kurakin2016adversarial} was introduced for leveraging finer gradient information by iteratively updating the perturbations with a smaller step size $\alpha$:
\begin{equation}
\label{IFGSM}
{\boldsymbol{x}_0'}=\boldsymbol{x},~~{\boldsymbol{x}_{k}'}={\boldsymbol{x}_{k-1}'}+\alpha\cdot\ \mathrm{sign}({\nabla_{{\boldsymbol{x}}}J(\boldsymbol{x}_{k-1}',y)}),
\end{equation}
where the intermediate perturbed image $\boldsymbol{x}_{k}'$ is projected onto a $\epsilon$-sphere around the original $\boldsymbol{x}$, to satisfy the $L_{\infty}$-norm constraint.
Note that I-FGSM constrains only the maximum change of individual image coordinates without considering the image-level accumulated difference.
For this reason, I-FGSM yields poor imperceptibility, especially in high-confidence settings (cf. Fig.~\ref{fig:visual}).

A generalization of I-FGSM to the $L_2$ norm can be achieved by changing the $\mathrm{sign}$ operation in the updating to:
\begin{equation}
\label{IFGSM_lp}
\frac{\nabla_{{\boldsymbol{x}}}J({\boldsymbol{x}_{k-1}'},y)}{{\|{\nabla_{{\boldsymbol{x}}}J({\boldsymbol{x}_{k-1}'},y)}\|}_2},
\end{equation}
where the projection is implemented by:
\begin{equation}
\label{IFGSM_lp_pro}
\boldsymbol{x}_{k}'=\boldsymbol{x}+\epsilon\frac{\boldsymbol{x}_{k}'-\boldsymbol{x}}{{\|\boldsymbol{x}_{k}'-\boldsymbol{x}\|}_2}.
\end{equation}
A recent method called the Decoupled Direction and Norm (DDN)~\cite{rony2019decoupling}, which is based on the $L_2$ norm-based I-FGSM, yielded the best performance (smallest $L_2$ norm) in the untargeted track of NIPS 2018 Adversarial Vision Challenge~\cite{brendel2018adversarial}, with substantially fewer iterations than the conventional C\&W.
In DNN, the $\epsilon$ is designed to be adjustable in each iteration based on whether the perturbed image is adversarial or not, leading to a finer search for the minimal norm.
Our second approach, PerC-AL, follows a similar strategy as DDN to improve efficiency by decoupling the joint optimization.

\subsection{Adversarial examples beyond $L_{p}$ norms}
\label{PerceptionAE}
 
Our work is part of the current movement away from tight $L_{p}$ norms and towards conceptualization of image similarity in terms of semantics or perceptual properties.  
Research that defines similarity in terms of semantics, requires the adversarial image to have the same content as the original image from the point of view of the human viewer.
Some of the first work in this direction has explored geometric transformation~\cite{engstrom2017rotation,xiao2018spatially}, global color shift~\cite{afifi2019else,bhattad2020Unrestricted,hosseini2018semantic,Laidlaw2019functional}, and image filters~\cite{choi2017geo}.

Such approaches are interesting, but we do not pursue them here because they tend to be limited in their adversarial strength, due to the restricted size of the search space for possible adversarial image transformations.

Research that investigates similarity with respect to texture and structure~\cite{croce2019sparse,gragnaniello2019perceptual,luo2018towards,wong2019wasserstein,zhang2020smooth}, has focused on hiding perturbations in image regions with visual variation.
Such hiding can be achieved by either using existing structure-aware metrics~\cite{gragnaniello2019perceptual,wong2019wasserstein}, such as structural similarity (SSIM)~\cite{wang2004image} and Wasserstein distance~\cite{kantorovich1958space}, or directly allowing more perturbations in the high-variance image regions~\cite{croce2019sparse,luo2018towards,zhang2020smooth}.
All of these approaches share a common challenge: They have difficulties in dealing with smooth regions (e.g., sky, ground and artificial objects), which appear frequently in images taken in commonly occurring real-world settings (referred to as \emph{natural images}).
In contrast, our PerC perturbations are applicable in smooth regions in the case of saturated color. 
Our experiments show that PerC perturbations can be combined productively with a structure-based approach (see Section~\ref{sec:texture}).

\section{Proposed approaches}
\label{method}
In this section, we present two approaches to using perceptual color (PerC) distance for adversarial image perturbations. 
We focus on image-level accumulated perceptual color difference, i.e., the $L_{2}$ norm of the color distance vector, in which each component represents the perceptual color distance ($\Delta E_{00}$ in Eq.~(\ref{deltaE})) calculated for the corresponding image pixel.

\subsection{Perceptual color distance penalty (PerC-C\&W)}
\label{method_opt}
Our first approach, PerC-C\&W, adopts the joint optimization of the well-known C\&W, but replaces the original penalty on the $L_2$ norm with a new one based on perceptual color difference.
It can be formally expressed as:
\begin{equation}
\label{opt-CP}
\underset{\boldsymbol{w}}{\mathrm{minimize}}
~~{{\|\Delta E_{00}(\boldsymbol{x},\boldsymbol{x}')\|}_2}+\lambda f(\boldsymbol{x}'),
\end{equation}
where $\boldsymbol{w}$ is the new introduced variable as in the Eq.~(\ref{cw}) of C\&W.
Like the original C\&W, the optimization problem is solved by binary search over the constant $\lambda$.
By using the gradient information from perceptual color difference, perturbation updating is translated into a perceptually uniform color space.
Large RGB perturbations, which have a strong adversarial effect, remain hidden from the human eye, as will be shown in Section~\ref{sec:exp}.

\begin{algorithm}[t]
\caption{Perceptual Color Distance Alternating Loss (PerC-AL)}
\label{PerC-AL}
\algrenewcommand\algorithmicrequire{\textbf{Input:}}
\algrenewcommand\algorithmicensure{\textbf{Output:}}
\algorithmicrequire{\\$\boldsymbol{x}$: original image, $t$: target label, $K$: number of iterations\\
$\alpha_l$: step size in minimizing classification loss\\
$\alpha_c$: step size in minimizing perceptual color difference\\}

\algorithmicensure{ $\boldsymbol{x}'$: adversarial image}
\begin{algorithmic}[1]
\State Initialize $\boldsymbol{x}_0'\leftarrow \boldsymbol{x}$, $\boldsymbol{\delta}_0\leftarrow \boldsymbol{0}$
\For {$k\leftarrow 1$ to $K$}
\If{$\boldsymbol{x}_{k-1}'$ is not adversarial}
\State $\boldsymbol{g}\leftarrow -\nabla_{\boldsymbol{x}}J(\boldsymbol{x}_{k-1}',t)$
\State $\boldsymbol{g}\leftarrow\alpha_l\cdot{\frac{\boldsymbol{g}}{{\|\boldsymbol{g}\|}_2}}$
\State $\boldsymbol{\delta}_k\leftarrow\boldsymbol{\delta}_{k-1}+\boldsymbol{g}$\Comment{\parbox[t]{.4\linewidth}}{Update $\boldsymbol{\delta}$ in the direction of $\boldsymbol{g}$}
\Else
\State $C_2\leftarrow{-\|\Delta E_{00}(\boldsymbol{x},\boldsymbol{x}_{k-1}')\|}_2$
\State $\boldsymbol{g}_c\leftarrow\nabla_{\boldsymbol{x}}C_2$
\State $\boldsymbol{g}_c\leftarrow\alpha_c\cdot{\frac{\boldsymbol{g}_c}{{\|\boldsymbol{g}_c\|}_2}}$
\State $\boldsymbol{\delta}_k\leftarrow\boldsymbol{\delta}_{k-1}+\boldsymbol{g}_c$\Comment{\parbox[t]{.35\linewidth}}{Update $\boldsymbol{\delta}$ in the direction of $\boldsymbol{g}_c$}
\EndIf
\State $\boldsymbol{x}_{k}'\leftarrow \mathrm{clip}(\boldsymbol{x}+\boldsymbol{\delta}_k,0,1)$
\State $\boldsymbol{x}_{k}'\leftarrow\mathrm{quantize}(\boldsymbol{x}_{k}')$\Comment{Ensure $\boldsymbol{x}_{k}'$ is valid}
\EndFor
\State \Return $\boldsymbol{x}'\leftarrow\boldsymbol{x}_{k}'$ that is adversarial and has smallest $C_2$
\end{algorithmic}
\end{algorithm}

\subsection{Perceptual color distance alternating loss (PerC-AL)}
\label{method_pgd}
Although, Eq.~\ref{opt-CP} enjoys a concise expression, the joint optimization of PerC-C\&W faces difficulties in practice.
Adversarial training~\cite{kurakin2016adversarial,madry2017towards}, for example, presents challenges. 
The reason is that PerC-C\&W requires time-consuming binary search in order to find an optimal $\lambda$, which normally varies substantially among different images~\cite{rony2019decoupling}.
To address the inefficiency, we propose PerC-AL, which decouples the joint optimization by alternately updating the perturbations with respect to either classification loss or perceptual color difference.
Our strategy is inspired by DDN, which is basically a projected gradient descent (PGD) method with a  dynamic $L_2$-norm bound.
However, PerC-AL goes beyond this idea to alternate two gradient descents. 

The full PerC-AL method is described in Algorithm~\ref{PerC-AL}.
We start from an original image $\boldsymbol{x}$ with the perturbation $\boldsymbol{\delta}$ initialized as $\boldsymbol{0}$, and iteratively update it to create an adversarial image.
In each iteration, the perturbation is either enlarged to achieve stronger adversarial effect based on the gradients from the classification loss, or shrunk to minimize perceptual color differences.
These two operations are alternated based on whether the intermediate perturbed image $\boldsymbol{x}_{k}'$ is adversarial or not, leading to a finer search of a minimal perceptual color difference by repeatedly crossing the decision boundary.
To ensure the adversarial image is valid, the output is clipped into the range [0,1] and quantized into 255 levels (corresponding to 8-bit image encoding).

\section{Experiments}
\label{sec:exp}
In this section, we first provide a picture of the differences between RGB and PerC approaches (Section~\ref{sec:adv}).
Then, we carry out experiments that compare different approaches in terms of robustness (Section~\ref{sec:robust}) and transferability (Section~\ref{sec:trans}) by considering the case of high-confidence adversarial examples.
Finally, in Section~\ref{sec:texture}, we show that structural information can be elegantly integrated into our efficient decoupled approach, PerC-AL, for further improvement in the imperceptibility of images that contain areas with rich visual variation.

\subsection{Experimental setup}
\label{sec:setup}
\noindent\textbf{Dataset and Networks.}
Following recent work~\cite{dong2019evading,xiao2018spatially,zhang2020smooth}, we conduct our experiments on the development set (1000 RGB natural images with the size of $299\times299$) of the ImageNet-Compatible dataset\footnote{\url{https://github.com/tensorflow/cleverhans/tree/master/examples/nips17_adversarial_competition/dataset}.}. 
This dataset was introduced by the NIPS 2017 Competition on Adversarial Attacks and Defenses~\cite{kurakin2018adversarial} and consists of
6000 images labeled with 1000 ImageNet classes.
We choose this dataset because we would like to study imperceptibility under real-world conditions. 
In contrast, some other work~\cite{croce2019sparse,luo2018towards} on addressing imperceptibility mainly focuses on the tiny images from MNIST~\cite{lecun1998gradient} and CIFAR-10~\cite{krizhevsky2009learning}. 
As in the competition, the Inception V3~\cite{szegedy2016rethinking} model pre-trained on ImageNet is used as the target classifier.

\noindent\textbf{Baselines.}
Three well-known baselines, I\nobreakdash-FGSM~\cite{kurakin2016adversarial}, C\&W~\cite{carlini2017towards}, and the state-of-the-art DDN~\cite{rony2019decoupling}, are compared with our approaches.
Among them, I-FGSM targets minimum $L_{\infty}$ norm, while C\&W and DDN target minimum $L_2$ norm.
Note that I-FGSM is not designed for imperceptibility, but we consider it here for completeness.

\noindent\textbf{Parameters.}
I-FGSM is repeated multiple rounds with increased $L_{\infty}$-norm bound, where in each round, a large enough iteration budget (100 in our implementation) is specified with the step size $\alpha=1/255$.

C\&W and PerC-C\&W use the Adam optimizer~\cite{kingma2014adam} with a learning rate of 0.01 for updating the perturbations.
We impose a budget on the number of search steps used to find the optimal $\lambda$.
The initialization of $\lambda$ is particularly important for small budgets. 
We perform grid search for the initialization value of $\lambda$ over the range [0.01, 0.1, 1, 10, 100], and adopt the value that yields the smallest average perturbation size.
The selected initialization values are shown in Table~\ref{tab:para} of the appendix.

For DDN and PerC-AL, we decrease the step size ($\alpha$ in DDN and $\alpha_l$ in PerC-AL) that is used for updating the perturbations with respect to the classification loss from 1 to 0.01 with cosine annealing.
The $L_2$-norm constraint $\epsilon$ in DDN is initialized to 1 and adjusted iteratively by $\gamma=0.05$, as in the original work DDN~\cite{rony2019decoupling}.
The $\alpha_c$ in PerC-AL is gradually reduced from 0.5 to 0.05 with cosine annealing.

\noindent\textbf{Evaluation Protocol.}
We investigate a set of reasonable operating points, based on pre-defined budgets.
Note that our goal is to show the relative behavior of PerC vs. RGB approaches.
For this purpose, we only need to create a fair comparison, and it is not necessary to drive all approaches to an absolute optimum.
For each image, an approach is considered successful if the perturbed image can achieve adversarial effect with the given budget.
Specifically, I\nobreakdash-FGSM requires varied repetitions for different images.
For C\&W and PerC-C\&W, the budget refers to N(search steps) $\times$ N(iterations of gradient descent).
We apply relatively high budget ($9\times1000$), and are also interested in lower budgets ($5\times200$ and $3\times100$), which are more directly comparable with more efficient approaches, namely, DDN and PerC-AL.
We test DDN and our PerC-AL with three different iteration budgets (100, 300 and 1000), adopted from the original work~\cite{rony2019decoupling}.

Adversarial strength is evaluated by the success rate, i.e., the proportion of successful cases over the whole dataset.
The averaged perturbation size over all successful images is reported. It is measured in terms of the $L_2$ and $L_{\infty}$ norm in RGB space ($\overline{L_2}$ and $\overline{L_{\infty}}$) and also in terms of image-level accumulated perceptual color difference ($\overline{C_2}$).

\begin{table}[t]
\newcommand{\tabincell}[2]{\begin{tabular}{@{}#1@{}}#2\end{tabular}}
\begin{center}
\resizebox{\columnwidth}{!}{
\begin{tabular}{l|c|c|ccc}
\toprule[1pt]
 \multirow{2}{*}{Approach}&\multirow{2}{*}{Budget}&Success&\multicolumn{3}{c}{Perturbation Size}\\
 &&Rate (\%)&$\overline{L_2}$&$\overline{L_{\infty}}$&$\overline{C_2}$
\\
\midrule[1pt]
I-FGSM~\cite{kurakin2016adversarial}
       &-&100.0&2.51&1.59&317.96\\
            \hline
\multirow{3}{*}{C\&W~\cite{carlini2017towards}}
        &3$\times$100&100.0&1.32&8.84&159.85\\
        &5$\times$200&100.0&1.09&8.20&132.86\\
        &9$\times$1000&100.0&0.92&8.45&114.36\\
            \hline
\multirow{3}{*}{PerC-C\&W (ours)}
        &3$\times$100&100.0&2.77&14.29&150.44\\
         &5$\times$200&100.0&1.48&12.06&83.93\\
         &9$\times$1000&100.0&1.22&15.57&67.79\\
            \hline            
\multirow{3}{*}{DDN~\cite{rony2019decoupling}}
        &100&100.0&1.00&7.84&136.11\\
        &300&100.0&0.88&7.58&120.12\\
       &1000&100.0&0.82&7.62&111.65\\
            \hline

\multirow{3}{*}{PerC-AL (ours)}
    &100&100.0&1.30&11.98&69.49\\
    &300&100.0&1.17&13.97&61.21\\
    &1000&100.0&1.13&17.04&57.10\\
  \bottomrule[1pt]
\end{tabular}
    }
\end{center}
\caption{Success rates and perturbation sizes on the 1000 images from the ImageNet-Compatible dataset, with varied budgets in the targeted setting.
Perturbation size is quantified in terms of $L_2$ and $L_{\infty}$ norms of the perturbations in RGB space ($\overline{L_2}$ and $\overline{L_{\infty}}$) and also in terms of image-level accumulated perceptual color difference ($\overline{C_2}$).
Note that C\&W and PerC-C\&W actually need more (here, 5$\times$) iterations to find the optimal initialization of $\lambda$.
The budget for I-FGSM varies on different images.}
\vspace{-0.3cm}

\label{tab:quan}
\end{table}

\subsection{Adversarial strength and imperceptibility}
\label{sec:adv}
In this section, we investigate the adversarial strength and imperceptibility of the perturbed images generated by different approaches in a white-box scenario, where the full information of the network is accessible.

\subsubsection{Sufficient-confidence adversarial examples}
\label{sec:just}
We first present, in Table~\ref{tab:quan}, a comparison demonstrating how PerC approaches relax $L_p$ norms.
Our comparison uses adversarial examples created under a commonly used condition where the aim is to achieve a just sufficient adversarial effect.
\emph{Sufficient-confidence adversarial examples} just cross the decision boundary without pursuing a higher confidence score for the adversarial label.
As expected, all approaches achieve 100\% success rate and the resulting perturbation size gets smaller as the budget increases.

Table~\ref{tab:quan}, which reports the targeted results, confirms that 
PerC approaches, PerC-C\&W and PerC-AL, show the behavior they are designed for, i.e., decreasing the average accumulated perceptual color difference $\overline{C_2}$.
More importantly, PerC approaches do this without tightly constraining the $L_p$ norms in RGB space as the other approaches do, as reflected by $\overline{L_2}$ and $\overline{L_{\infty}}$.
Moreover, PerC-AL achieves lower $\overline{C_2}$ than PerC-C\&W (57.10 vs. 67.79) with notably fewer iterations. 
For comparison, we provide $\overline{C_2}$ for the RGB approaches. 
The untargeted results follow a similar pattern and can be found in Table~\ref{tab:quan_1}
of the appendix.

\begin{table}[t]
\newcommand{\tabincell}[2]{\begin{tabular}{@{}#1@{}}#2\end{tabular}}
\begin{center}
\resizebox{\columnwidth}{!}{
\begin{tabular}{l|cccc}
\toprule[1pt]
\multirow{2}{*}{Approach}&\multicolumn{2}{c}{$\kappa=20$}&\multicolumn{2}{c}{$\kappa=40$}\\
&Suc. (\%)&$\overline{C_2}$&Suc. (\%)&$\overline{C_2}$\\
\midrule[1pt]
I-FGSM~\cite{kurakin2016adversarial}&100.0&375.74&99.9&576.06\\
\hline
C\&W~\cite{carlini2017towards}&100.0&159.00&100.0&241.92\\
\hline
DDN~\cite{rony2019decoupling}&100.0&150.68&98.1&238.37\\
\hline
PerC-C\&W (ours)&100.0&90.86&100.0&136.22\\
\hline
PerC-AL (ours)&100.0&75.43&100.0&115.17\\
  \bottomrule[1pt]
\end{tabular}
    }
        
\end{center}
\caption{Evaluation of the success rate and perceptual color difference achieved by different approaches in high-confidence settings.}
\label{tab:high_conf}
\vspace{-0.2cm}

\end{table}

\begin{figure*}[t]
\begin{center}
  \includegraphics[width=0.9\textwidth]{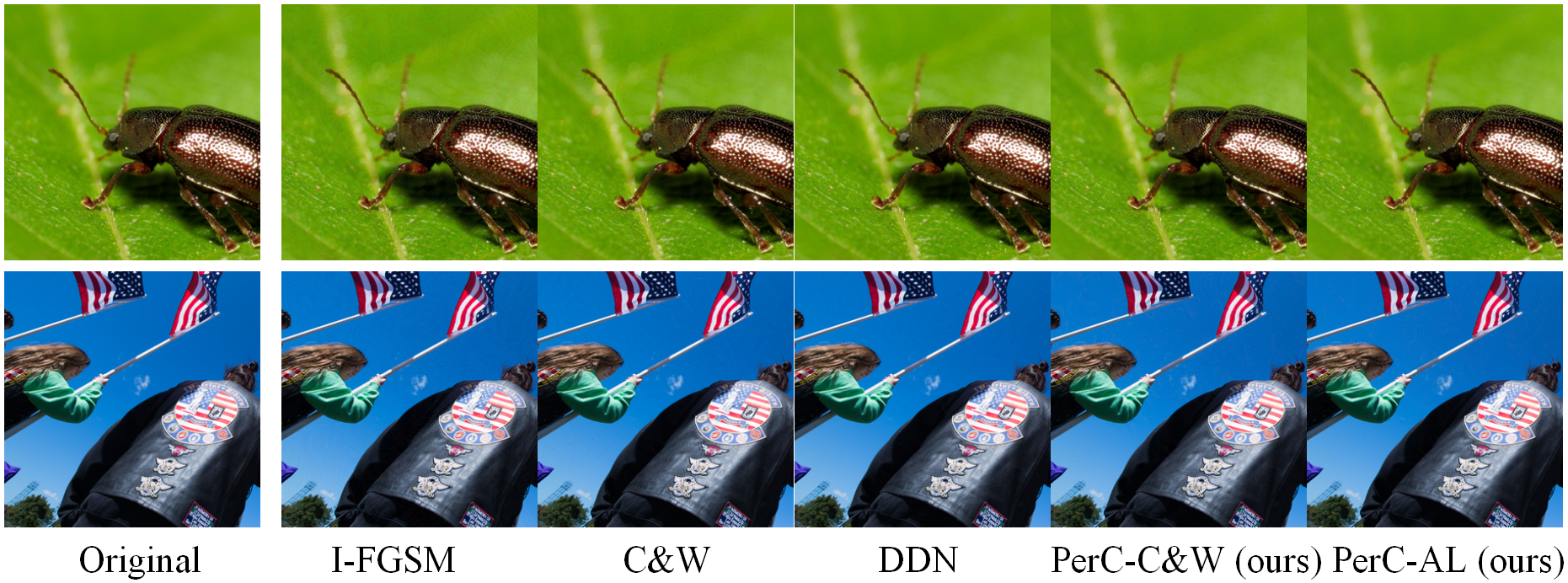}
\end{center}
\vspace{-0.2cm}
   \caption{Examples of adversarial images generated by five different approaches with high confidence level $\kappa=40$}
\label{fig:visual}
\vspace{-0.2cm}

\end{figure*}

\subsubsection{High-confidence adversarial examples}
\label{sec:high}
In order to gain deeper insight into the performance of our approaches, we investigate adversarial examples that have a high confidence score for the adversarial label.
High confidence was initially investigated by~\cite{carlini2017towards} in order to achieve more transferable adversarial examples, and also been explored in the ``Unrestricted Adversarial Examples'' contest~\cite{brown2018unrestricted}.
In the untargeted setting, an approach is regarded as successful only if the logit with respect to the original class becomes lower than the maximum of the other logits by a pre-defined margin $\kappa$.
For C\&W and PerC-C\&W, this requirement can be directly implemented by specifying the factor $\kappa$ in Eq.~(\ref{eq:cw_untar}).
For I\nobreakdash-FGSM, DDN and PerC-AL, this can be achieved by running the iterations until the required logit difference is satisfied.
For this experiment, we adopt the settings generating the smallest perturbations for each approach in Section~\ref{sec:just}.

Fig.~\ref{fig:visual} shows some adversarial examples generated by different approaches at $\kappa=40$.
The images produced by our PerC approaches look more visually acceptable than those of the other approaches (best viewed on screen).
The good visual appearance of the PerC examples is consistent with their low averaged aggregated perceptual color difference, $\overline{C_2}$, as seen in Table~\ref{tab:high_conf}, which shows both $\kappa=40$ and $\kappa=20$ values.
The challenge of the high-confidence setting is seen in the success rates, which are not longer perfect for all conditions.
 
\begin{figure}[t]
\begin{center}
  \includegraphics[width=\columnwidth]{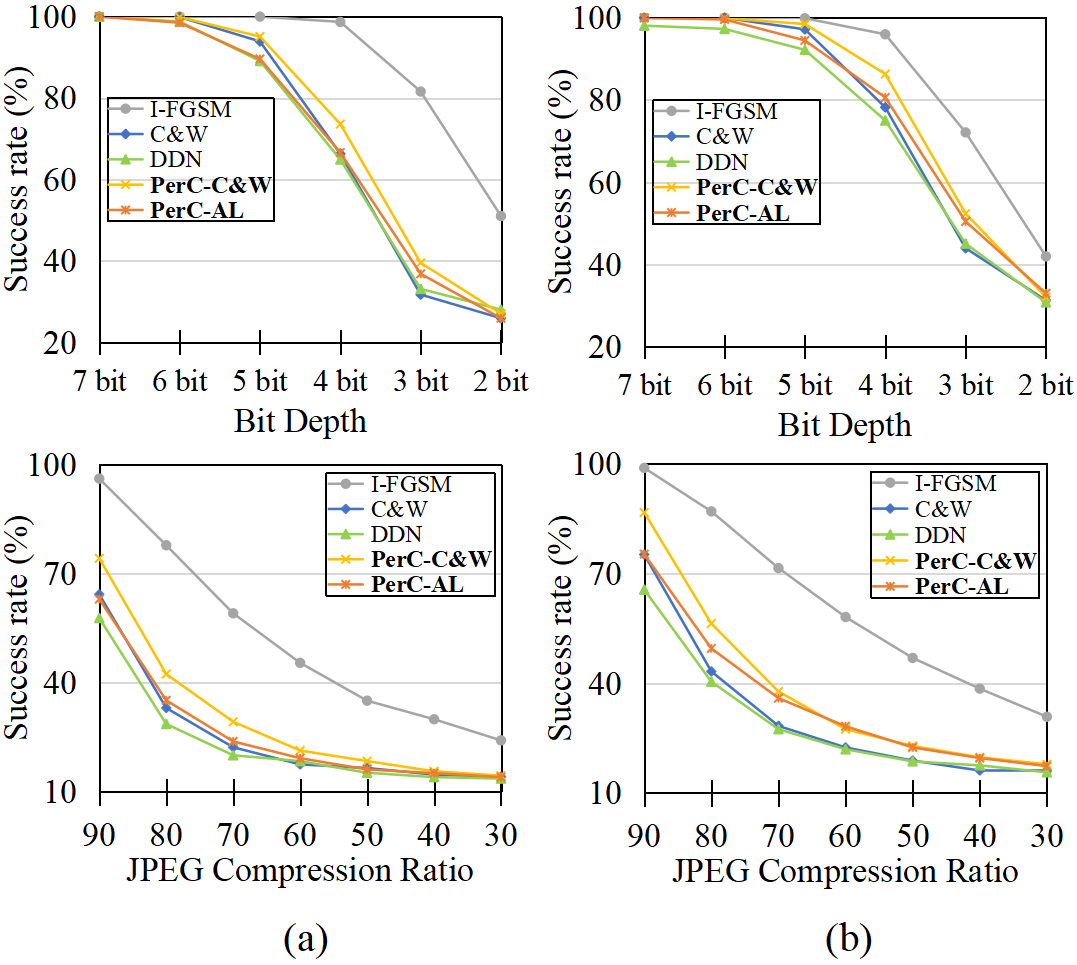}
\end{center}
\vspace{-0.2cm}
   \caption{Evaluation of robustness of high-confidence adversarial examples at (a) $\kappa=20$ and (b) $\kappa=40$, against two types of image transformations: JPEG compression (top row) and bit-depth reduction (bottom row).}
\label{fig:rob}
\vspace{-0.2cm}
\end{figure}

\subsection{Robustness}
\label{sec:robust}
In order to gain additional practical insight, we test the robustness of the adversarial examples against two commonly studied image transformation-based defense methods, i.e., JPEG compression~\cite{das2018shield,dong2019evading,dziugaite2016study,guo2017countering} and bit-depth reduction~\cite{guo2017countering,he2017adversarial,xu2017feature}.

The results are shown in Fig.~\ref{fig:rob}.
Overall, increasing $\kappa$ from 20 to 40 leads to improved robustness.
For a specific $\kappa$, unsurprisingly, I-FGSM outperforms other approaches since it greedily perturbs all the pixels, but at the cost of worse image quality (see Fig.~\ref{fig:visual}).
Among the other four approaches that target minimal image-level accumulated image difference with sparse perturbations, the best results are consistently achieved by either our PerC-C\&W or PerC-AL.
Specifically, PerC-C\&W outperforms the original C\&W in all cases, while PerC-AL consistently outperforms DDN.
Recall that our PerC approaches cause fewer visual distortions, as shown in Fig.~\ref{fig:visual}, contributing to imperceptibility.

\subsection{Transferability}
\label{sec:trans}
Existing research~\cite{liu2016delving,tramer2017ensemble} has demonstrated that the adversarial effect of images optimized with respect to a specific network may
transfer to another network.
We test the transferability of different approaches from the original Inception V3 to other three pre-trained networks, namely, GoogLeNet~\cite{szegedy2016rethinking}, ResNet-152~\cite{he2016deep}, and VGG-16~\cite{simonyan2014very}.
Specifically, an untargeted adversarial example generated for the original model is regarded to be transferable to a new model if it can also induce misclassification of that model.

We report results on a subset of our data containing images that all four models originally classify correctly.
Table~\ref{tab:trans} reports the success rates under transferability on these images (767 in total).
I-FGSM again outperforms the other approaches, but uses excessive perturbations (and for this reason is shown in italics).
Among the other approaches, we can observe that the best results are always achieved by one of our two PerC approaches\footnote{Results in Table~\ref{tab:trans} are different from those in previous arXiv version due to changed normalization implementations, while the claims still hold.}.
        

\begin{table}[t]
\newcommand{\tabincell}[2]{\begin{tabular}{@{}#1@{}}#2\end{tabular}}
\renewcommand{\arraystretch}{1}
\begin{center}
\resizebox{\columnwidth}{!}{
\begin{tabular}{l|cc|cc|cc}
\toprule[1pt]
&\multicolumn{2}{c|}{GoogLeNet}&\multicolumn{2}{c|}{VGG-16}&\multicolumn{2}{c}{ResNet-152}\\
&$\kappa=20$&$\kappa=40$&$\kappa=20$&$\kappa=40$&$\kappa=20$&$\kappa=40$\\
\midrule[1pt]
I-FGSM~\cite{kurakin2016adversarial}&\textit{4.2}&\textit{6.3}&\textit{5.6}&\textit{10.6} &\textit{2.4}&\textit{4.2}\\
\hline
C\&W~\cite{carlini2017towards}&2.5&3.1&3.0&5.1&0.9& 1.7\\
\hline
DDN~\cite{rony2019decoupling}&2.1&3.1&3.3&5.7&\underline{1.2}& 2.4\\
\hline
PerC-C\&W (ours)&\underline{2.9}&\underline{4.7}&3.9&6.9&\underline{1.2}&\underline{2.5}\\
\hline
PerC-AL (ours)&2.6&4.3&\underline{4.8}&\underline{7.2}&\underline{1.2}&2.4\\
  \bottomrule[1pt]
\end{tabular}
    }
        
\end{center}
\vspace{-0.2cm}
\caption{Success rates of adversarial examples at two high confidence levels $\kappa=20$ and $\kappa=40$ in the transfer scenario, from the source model Inception V3 to three others.}
\label{tab:trans}
\vspace{-0.2cm}
\end{table}

\subsection{Assembling structural information}
\label{sec:texture}
\begin{figure}[t]
\begin{center}
  \includegraphics[width=\columnwidth]{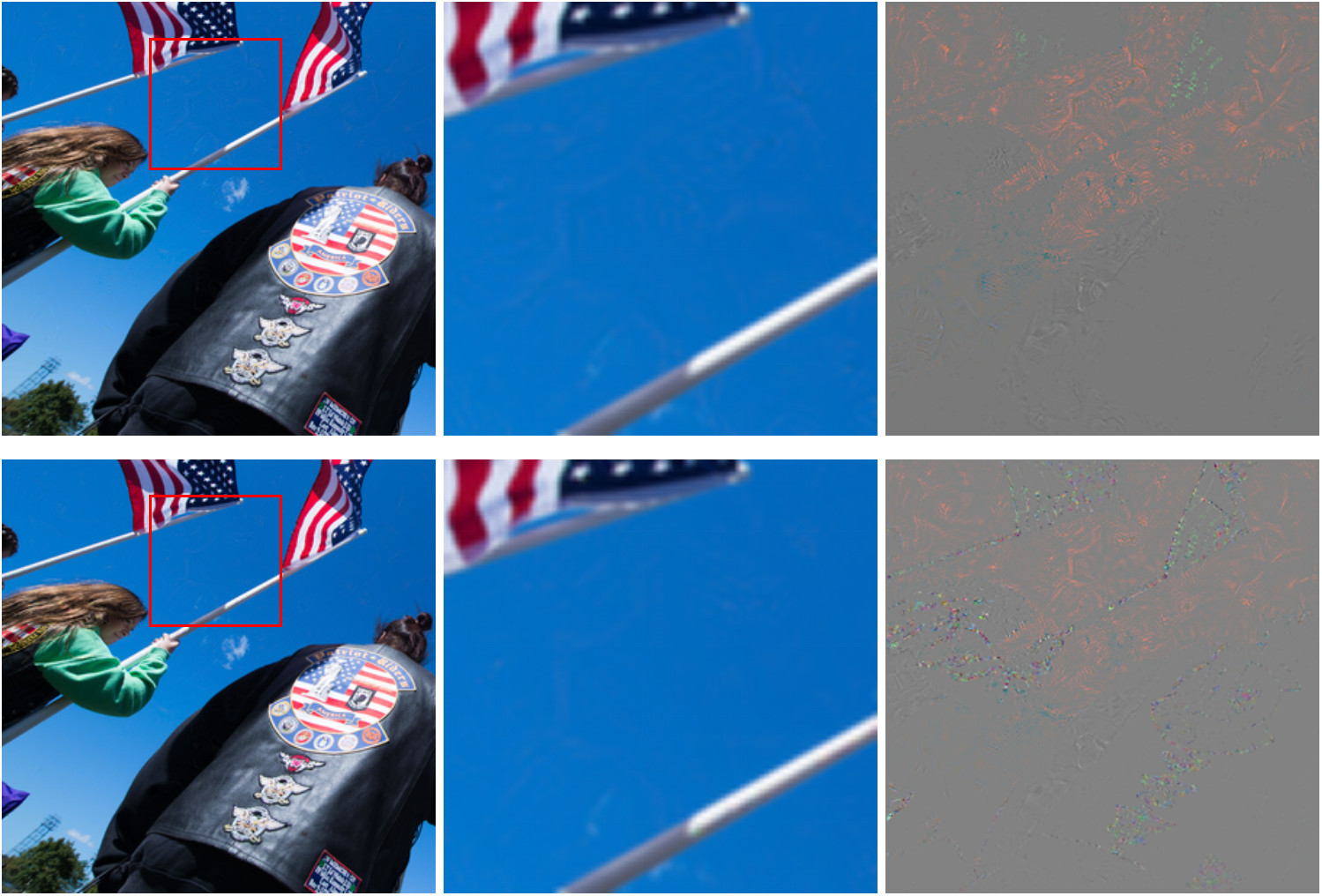}
\end{center}
\vspace{-0.4cm}
   \caption{Adversarial examples of an image at $\kappa=40$. Top: Generated by PerC-AL (Algorithm~\ref{PerC-AL}). Bottom: Generated by PerC-AL plus structure (Algorithm~\ref{PerC-AL} plus Eq.~(\ref{PerC-AL-t})).}
\label{fig:texture}
\vspace{-0.4cm}
\end{figure}

We explore the possibility of assembling structural information for further improving imperceptibility without impacting adversarial strength.
Specifically, we introduce a texture complexity vector $\boldsymbol{\sigma}$, which has the same size as the image, as a weighting term into our PerC-AL framework.
Following existing work~\cite{croce2019sparse,luo2018towards} on addressing imperceptibility with respect to image structure, this vector is obtained by calculating the standard deviation of the pixel values in each $3\times3$ square per channel.
The components with top 5\% highest values in the map are clipped for stability and the map is normalized into the range [0,1] before use.
Concretely, step 8 in Algorithm~\ref{PerC-AL} is adjusted to:
\begin{equation}
\label{PerC-AL-t}
C_2\leftarrow -{\|(\boldsymbol{1}-\boldsymbol{\sigma})\cdot \Delta E_{00}(\boldsymbol{x},\boldsymbol{x}_{k-1}')\|}_2,
\end{equation}
where $C_2$ also becomes sensitive to image differences in terms of local visual variation.
As shown in Fig.~\ref{fig:texture}, with the help of additional structural information, perturbations in the smooth regions are suppressed, while more changes, which are barely perceptible, are triggered in the area with rich visual variation.
It is worthwhile for future work to investigate this combined approach in more detail.

\section{Conclusion and Outlook}
This paper has demonstrated the usefulness of perceptual color distance for creating large yet imperceptible adversarial image perturbations.
We have proposed two approaches for creating adversarial images, PerC-C\&W and PerC-AL.
Our experimental investigation of these approaches shows that perceptual color distance is able to improve imperceptibility, especially in smooth, saturated regions.
We show that these approaches have perturbations with larger RGB $L_p$ norms than approaches that perturb directly in RGB space.
This effect translates into adversarial strength, i.e., the ability of the perturbations to fool a classifier.

Our work has made a contribution to recent work that seeks to create adversarial images that are imperceptible to the eye of the human observer.
This work has been carried out in the area of security~\cite{carlini2017towards,eykholt2017robust,gragnaniello2019perceptual,kurakin2016adversarial,papernot2016limitations} (defend inference of a legitimate classifier) and privacy~\cite{choi2017geo,liu2019s,mirjalili2018gender,oh2017adversarial} (prevent inference of an illegitimate classifier).
In the security area, imperceptible perturbations can mean that adversarial images can poison the training data without being noticed by human annotators.
In the privacy area, imperceptible perturbations mean wider acceptance of the use of adversarial images to protect against classification attacks.

In the future, we will continue to consider perceptual color in adversarial images from both the privacy and the security angle. 
Our first direction will be related to the fact that neither conventional RGB perturbations nor PerC perturbations perform well in smooth regions with low saturation. 
We would like to develop techniques that can make perturbations imperceptible, or unnecessary, in such regions.
Our future work will also look at model robustness specifically against our PerC adversaries.
On one hand, adversarial training on images perturbed by PerC is worth exploring to complement current research on $L_p$ robustness.
On the other hand, it would be interesting to investigate ways to detect whether PerC has been applied to an image, or design countering methods that can mitigate PerC-based perturbations by, for example,  applying bit-depth reduction directly in perceptual color space.

\section*{Acknowledgement}
This work was carried out on the Dutch national e-infrastructure with the support of SURF Cooperative.
\clearpage
\newpage

{\small
\bibliographystyle{ieee_fullname}
\bibliography{ref}

\begin{thebibliography}{10}\itemsep=-1pt

\bibitem{ciede2000}
{ISO/CIE 11664-6:2014(E) Colourimetry-Part 6: CIEDE2000 Colour difference
  Formula}.

\bibitem{afifi2019else}
Mahmoud Afifi and Michael~S Brown.
\newblock What else can fool deep learning? {A}ddressing color constancy errors
  on deep neural network performance.
\newblock In {\em ICCV}, pages 243--252, 2019.

\bibitem{athalye2018synthesizing}
Anish Athalye, Logan Engstrom, Andrew Ilyas, and Kevin Kwok.
\newblock Synthesizing robust adversarial examples.
\newblock In {\em ICML}, pages 284--293, 2018.

\bibitem{bhattad2020Unrestricted}
Anand Bhattad, Min~Jin Chong, Kaizhao Liang, Bo Li, and David~A Forsyth.
\newblock Unrestricted adversarial examples via semantic manipulation.
\newblock In {\em ICLR}, 2020.

\bibitem{brendel2018adversarial}
Wieland Brendel, Jonas Rauber, Alexey Kurakin, Nicolas Papernot, Behar Veliqi,
  Marcel Salath{\'e}, Sharada~P. Mohanty, and Matthias Bethge.
\newblock Adversarial vision challenge.
\newblock {\em arXiv preprint arXiv:1808.01976}, 2018.

\bibitem{brown2018unrestricted}
Tom~B Brown, Nicholas Carlini, Chiyuan Zhang, Catherine Olsson, Paul
  Christiano, and Ian Goodfellow.
\newblock Unrestricted adversarial examples.
\newblock {\em arXiv preprint arXiv:1809.08352}, 2018.

\bibitem{carlini2017towards}
Nicholas Carlini and David Wagner.
\newblock Towards evaluating the robustness of neural networks.
\newblock In {\em IEEE Symposium on Security and Privacy (S\&P)}, pages 39--57,
  2017.

\bibitem{cheng2001color}
Heng-Da Cheng, Xihua Jiang, Ying Sun, and Jingli Wang.
\newblock Color image segmentation: advances and prospects.
\newblock {\em Pattern Recognition}, 34(12):2259--2281, 2001.

\bibitem{choi2017geo}
Jaeyoung Choi, Martha Larson, Xinchao Li, Kevin Li, Gerald Friedland, and Alan
  Hanjalic.
\newblock The geo-privacy bonus of popular photo enhancements.
\newblock In {\em ICMR}, pages 84--92, 2017.

\bibitem{cohen2019certified}
Jeremy Cohen, Elan Rosenfeld, and Zico Kolter.
\newblock Certified adversarial robustness via randomized smoothing.
\newblock In {\em ICML}, pages 1310--1320, 2019.

\bibitem{croce2019sparse}
Francesco Croce and Matthias Hein.
\newblock Sparse and imperceivable adversarial attacks.
\newblock In {\em ICCV}, pages 4724--4732, 2019.

\bibitem{das2018shield}
Nilaksh Das, Madhuri Shanbhogue, Shang-Tse Chen, Fred Hohman, Siwei Li, Li
  Chen, Michael~E. Kounavis, and Duen~Horng Chau.
\newblock Shield: Fast, practical defense and vaccination for deep learning
  using {JPEG} compression.
\newblock In {\em SIGKDD}, pages 196--204, 2018.

\bibitem{dong2019evading}
Yinpeng Dong, Tianyu Pang, Hang Su, and Jun Zhu.
\newblock Evading defenses to transferable adversarial examples by
  translation-invariant attacks.
\newblock In {\em CVPR}, pages 4312--4321, 2019.

\bibitem{dziugaite2016study}
Gintare~Karolina Dziugaite, Zoubin Ghahramani, and Daniel~M. Roy.
\newblock A study of the effect of {JPG} compression on adversarial images.
\newblock {\em arXiv preprint arXiv:1608.00853}, 2016.

\bibitem{engstrom2017rotation}
Logan Engstrom, Brandon Tran, Dimitris Tsipras, Ludwig Schmidt, and Aleksander
  Madry.
\newblock A rotation and a translation suffice: Fooling {CNN}s with simple
  transformations.
\newblock In {\em NIPS 2017 Workshop on Machine Learning and Computer
  Security}, 2017.

\bibitem{gragnaniello2019perceptual}
Gragnaniello et al.
\newblock Perceptual quality-preserving black-box attack against deep learning
  image classifiers.
\newblock {\em IEEE Transactions on Information Forensics and Security (TIFS)},
  2019.

\bibitem{eykholt2017robust}
Kevin Eykholt, Ivan Evtimov, Earlence Fernandes, Bo Li, Amir Rahmati, Chaowei
  Xiao, Atul Prakash, Tadayoshi Kohno, and Dawn Song.
\newblock Robust physical-world attacks on deep learning models.
\newblock In {\em CVPR}, pages 1625--1634, 2018.

\bibitem{gatys2016preserving}
Leon~A. Gatys, Matthias Bethge, Aaron Hertzmann, and Eli Shechtman.
\newblock Preserving color in neural artistic style transfer.
\newblock {\em arXiv preprint arXiv:1606.05897}, 2016.

\bibitem{goodfellow2014explaining}
Ian Goodfellow, Jonathon Shlens, and Christian Szegedy.
\newblock Explaining and harnessing adversarial examples.
\newblock In {\em ICLR}, 2015.

\bibitem{guo2017countering}
Chuan Guo, Mayank Rana, Moustapha Cisse, and Laurens van~der Maaten.
\newblock Countering adversarial images using input transformations.
\newblock In {\em ICLR}, 2018.

\bibitem{he2016deep}
Kaiming He, Xiangyu Zhang, Shaoqing Ren, and Jian Sun.
\newblock Deep residual learning for image recognition.
\newblock In {\em CVPR}, pages 770--778, 2016.

\bibitem{he2017adversarial}
Warren He, James Wei, Xinyun Chen, Nicholas Carlini, and Dawn Song.
\newblock Adversarial example defense: Ensembles of weak defenses are not
  strong.
\newblock In {\em {USENIX} Workshop on Offensive Technologies}, 2017.

\bibitem{hosseini2018semantic}
Hossein Hosseini and Radha Poovendran.
\newblock Semantic adversarial examples.
\newblock In {\em CVPR Workshops}, pages 1614--1619, 2018.

\bibitem{joshi2019semantic}
Ameya Joshi, Amitangshu Mukherjee, Soumik Sarkar, and Chinmay Hegde.
\newblock Semantic adversarial attacks: Parametric transformations that fool
  deep classifiers.
\newblock In {\em ICCV}, pages 4773--4783, 2019.

\bibitem{kantorovich1958space}
Leonid~Vasilevich Kantorovich and Gennady~S Rubinstein.
\newblock On a space of completely additive functions.
\newblock {\em Vestnik Leningrad. Univ}, 13(7):52--59, 1958.

\bibitem{khan2012color}
Fahad~Shahbaz Khan, Rao~Muhammad Anwer, Joost Van De~Weijer, Andrew~D.
  Bagdanov, Maria Vanrell, and Antonio~M. Lopez.
\newblock Color attributes for object detection.
\newblock In {\em CVPR}, pages 3306--3313, 2012.

\bibitem{kingma2014adam}
Diederik~P. Kingma and Jimmy Ba.
\newblock Adam: A method for stochastic optimization.
\newblock In {\em ICLR}, 2014.

\bibitem{krizhevsky2009learning}
Alex Krizhevsky.
\newblock Learning multiple layers of features from tiny images.
\newblock {\em Master’s thesis, Department of Computer Science, University of
  Toronto}, 2009.

\bibitem{kurakin2016adversarial}
Alexey Kurakin, Ian Goodfellow, and Samy Bengio.
\newblock Adversarial examples in the physical world.
\newblock In {\em ICLR}, 2017.

\bibitem{kurakin2018adversarial}
Alexey Kurakin, Ian Goodfellow, Samy Bengio, Yinpeng Dong, Fangzhou Liao, Ming
  Liang, Tianyu Pang, Jun Zhu, Xiaolin Hu, Cihang Xie, et~al.
\newblock Adversarial attacks and defences competition.
\newblock In {\em The NIPS'17 Competition: Building Intelligent Systems}, pages
  195--231. 2018.

\bibitem{Laidlaw2019functional}
Cassidy Laidlaw and Soheil Feizi.
\newblock Functional adversarial attacks.
\newblock In {\em NeurIPS}, 2019.

\bibitem{lecun1998gradient}
Yann LeCun, L{\'e}on Bottou, Yoshua Bengio, Patrick Haffner, et~al.
\newblock Gradient-based learning applied to document recognition.
\newblock {\em Proceedings of the IEEE}, 86(11):2278--2324, 1998.

\bibitem{liu1989limited}
Dong~C Liu and Jorge Nocedal.
\newblock On the limited memory {BFGS} method for large scale optimization.
\newblock {\em Mathematical programming}, 45(1-3):503--528, 1989.

\bibitem{liu2010colorization}
Shuaicheng Liu, Michael~S Brown, Seon~Joo Kim, and Yu-Wing Tai.
\newblock Colorization for single image super resolution.
\newblock In {\em ECCV}, pages 323--336, 2010.

\bibitem{liu2016delving}
Yanpei Liu, Xinyun Chen, Chang Liu, and Dawn Song.
\newblock Delving into transferable adversarial examples and black-box attacks.
\newblock In {\em ICLR}, 2017.

\bibitem{liu2019s}
Zhuoran Liu, Zhengyu Zhao, and Martha Larson.
\newblock Who's afraid of adversarial queries? {T}he impact of image
  modifications on content-based image retrieval.
\newblock In {\em ICMR}, pages 306--314, 2019.

\bibitem{luo2018towards}
Bo Luo, Yannan Liu, Lingxiao Wei, and Qiang Xu.
\newblock Towards imperceptible and robust adversarial example attacks against
  neural networks.
\newblock In {\em AAAI}, 2018.

\bibitem{luo2001development}
Ming~Ronnier Luo, Guihua Cui, and B. Rigg.
\newblock The development of the {CIE} 2000 colour-difference formula:
  {CIEDE}2000.
\newblock {\em Color Research and Application}, 26(5):340--350, 2001.

\bibitem{madry2017towards}
Aleksander Madry, Aleksandar Makelov, Ludwig Schmidt, Dimitris Tsipras, and
  Adrian Vladu.
\newblock Towards deep learning models resistant to adversarial attacks.
\newblock In {\em ICLR}, 2018.

\bibitem{mirjalili2018gender}
Vahid Mirjalili, Sebastian Raschka, and Arun Ross.
\newblock Gender privacy: An ensemble of semi adversarial networks for
  confounding arbitrary gender classifiers.
\newblock In {\em IEEE International Conference on Biometrics Theory,
  Applications and Systems (BTAS)}, pages 1--10, 2018.

\bibitem{moosavi2016deepfool}
Seyed-Mohsen Moosavi-Dezfooli, Alhussein Fawzi, and Pascal Frossard.
\newblock {DeepFool}: a simple and accurate method to fool deep neural
  networks.
\newblock In {\em CVPR}, pages 2574--2582, 2016.

\bibitem{oh2017adversarial}
Seong~Joon Oh, Mario Fritz, and Bernt Schiele.
\newblock Adversarial image perturbation for privacy protection a game theory
  perspective.
\newblock In {\em ICCV}, pages 1491--1500, 2017.

\bibitem{papernot2016limitations}
Nicolas Papernot, Patrick McDaniel, Somesh Jha, Matt Fredrikson, Z~Berkay
  Celik, and Ananthram Swami.
\newblock The limitations of deep learning in adversarial settings.
\newblock In {\em IEEE European Symposium on Security and Privacy (EuroS\&P)},
  pages 372--387, 2016.

\bibitem{rony2019decoupling}
J{\'e}r{\^o}me Rony, Luiz~G. Hafemann, Luiz~S. Oliveira, Ismail~Ben Ayed,
  Robert Sabourin, and Eric Granger.
\newblock Decoupling direction and norm for efficient gradient-based {$L_2$}
  adversarial attacks and defenses.
\newblock In {\em CVPR}, pages 4322--4330, 2019.

\bibitem{sharif2018suitability}
Mahmood Sharif, Lujo Bauer, and Michael~K. Reiter.
\newblock On the suitability of {$L_p$}-norms for creating and preventing
  adversarial examples.
\newblock In {\em CVPR Workshops}, pages 1605--1613, 2018.

\bibitem{sharif2019adversarial}
Mahmood Sharif, Sruti Bhagavatula, Lujo Bauer, and Michael~K. Reiter.
\newblock A general framework for adversarial examples with objectives.
\newblock {\em ACM Transactions on Privacy and Security (TOPS)}, 2019.

\bibitem{simonyan2014very}
Karen Simonyan and Andrew Zisserman.
\newblock Very deep convolutional networks for large-scale image recognition.
\newblock In {\em ICLR}, 2015.

\bibitem{szegedy2016rethinking}
Christian Szegedy, Vincent Vanhoucke, Sergey Ioffe, Jon Shlens, and Zbigniew
  Wojna.
\newblock Rethinking the {Inception} architecture for computer vision.
\newblock In {\em CVPR}, pages 2818--2826, 2016.

\bibitem{szegedy2013intriguing}
Christian Szegedy, Wojciech Zaremba, Ilya Sutskever, Joan Bruna, Dumitru Erhan,
  Ian Goodfellow, and Rob Fergus.
\newblock Intriguing properties of neural networks.
\newblock In {\em ICLR}, 2014.

\bibitem{taigman2016unsupervised}
Yaniv Taigman, Adam Polyak, and Lior Wolf.
\newblock Unsupervised cross-domain image generation.
\newblock In {\em ICLR}, 2017.

\bibitem{tramer2017ensemble}
Florian Tram{\`e}r, Alexey Kurakin, Nicolas Papernot, Ian Goodfellow, Dan
  Boneh, and Patrick McDaniel.
\newblock Ensemble adversarial training: Attacks and defenses.
\newblock In {\em ICLR}, 2018.

\bibitem{van2009evaluating}
Koen Van De~Sande, Theo Gevers, and Cees Snoek.
\newblock Evaluating color descriptors for object and scene recognition.
\newblock {\em IEEE Transactions on Pattern Analysis and Machine Intelligence
  (TPAMI)}, 32(9):1582--1596, 2009.

\bibitem{wang2004image}
Zhou Wang, Alan~C. Bovik, Hamid~R. Sheikh, and Eero~P. Simoncelli.
\newblock Image quality assessment: from error visibility to structural
  similarity.
\newblock {\em IEEE Transactions On Image Processing (TIP)}, 13(4):600--612,
  2004.

\bibitem{wong2018provable}
Eric Wong and Zico Kolter.
\newblock Provable defenses against adversarial examples via the convex outer
  adversarial polytope.
\newblock In {\em ICML}, pages 5283--5292, 2018.

\bibitem{wong2019wasserstein}
Eric Wong, Frank Schmidt, and Zico Kolter.
\newblock Wasserstein adversarial examples via projected sinkhorn iterations.
\newblock In {\em ICML}, pages 6808--6817, 2019.

\bibitem{xiao2018spatially}
Chaowei Xiao, Jun-Yan Zhu, Bo Li, Warren He, Mingyan Liu, and Dawn Song.
\newblock Spatially transformed adversarial examples.
\newblock In {\em ICLR}, 2018.

\bibitem{xu2017feature}
Weilin Xu, David Evans, and Yanjun Qi.
\newblock Feature squeezing: Detecting adversarial examples in deep neural
  networks.
\newblock In {\em Network and Distributed Systems Security Symposium (NDSS)},
  2018.

\bibitem{yang2012color}
Yang Yang, Jun Ming, and Nenghai Yu.
\newblock Color image quality assessment based on {CIEDE}2000.
\newblock {\em Advances in Multimedia}, 2012:11, 2012.

\bibitem{zhang2020smooth}
Hanwei Zhang, Yannis Avrithis, Teddy Furon, and Laurent Amsaleg.
\newblock Smooth adversarial examples.
\newblock {\em EURASIP Journal on Information Security (JIS)}, 2020.

\end{thebibliography}
}
\clearpage
\newpage
\section*{Appendix}

\begin{table}[h]
\newcommand{\tabincell}[2]{\begin{tabular}{@{}#1@{}}#2\end{tabular}}
\begin{center}
\resizebox{\columnwidth}{!}{
\begin{tabular}{c|c|cc}
\toprule[1pt]
 \multirow{2}{*}{Approach}&\multirow{2}{*}{Budget}&\multicolumn{2}{c}{$\lambda$}\\
 &&Targeted&Untargeted
\\
\midrule[1pt]
\multirow{3}{*}{C\&W~\cite{carlini2017towards}}
        &3$\times$100&1&0.1\\
         &5$\times$200&1&1\\
         &9$\times$1000&1&1\\
            \hline
\multirow{3}{*}{PerC-C\&W (ours)}
        &3$\times$100&10&100\\
         &5$\times$200&10&100\\
         &9$\times$1000&10&10\\            
  \bottomrule[1pt]
\end{tabular}
    }
        
\end{center}
\caption{Selected initializations of $\lambda$ via grid search.}
\label{tab:para}
\end{table}

\begin{table}[h]
\newcommand{\tabincell}[2]{\begin{tabular}{@{}#1@{}}#2\end{tabular}}
\begin{center}
\resizebox{\columnwidth}{!}{
\begin{tabular}{l|c|c|ccc}
\toprule[1pt]
 \multirow{2}{*}{Approach}&\multirow{2}{*}{Budget}&Success&\multicolumn{3}{c}{Perturbation Size}\\
 &&Rate (\%)&$\overline{L_2}$&$\overline{L_{\infty}}$&$\overline{C_2}$
\\
\midrule[1pt]
I-FGSM~\cite{kurakin2016adversarial}
       &-&100.0&1.94&1.02&255.92\\
            \hline
\multirow{3}{*}{C\&W~\cite{carlini2017towards}}
        &3$\times$100&100.0&0.69&3.61&88.76\\
        &5$\times$200&100.0&0.45&3.79&59.88\\
        &9$\times$1000&100.0&0.41&3.74&54.17\\
            \hline
\multirow{3}{*}{PerC-C\&W (ours)}
        &3$\times$100&100.0&1.47&6.78&78.25\\
         &5$\times$200&100.0&0.90&6.71&51.35\\
         &9$\times$1000&100.0&0.56&6.58&33.00\\
            \hline            
\multirow{3}{*}{DDN~\cite{rony2019decoupling}}
        &100&100.0&0.35&4.03&49.43\\
        &300&100.0&0.33&4.08&47.58\\
       &1000&100.0&0.32&4.11&46.51\\
            \hline

\multirow{3}{*}{PerC-AL (ours)}
    &100&100.0&0.53&5.58&30.39\\
    &300&100.0&0.50&6.93&27.65\\
    &1000&100.0&0.51&8.92&26.62\\
  \bottomrule[1pt]
\end{tabular}
    }
\end{center}
\caption{Success rates and perturbation sizes on the 1000 images from the ImageNet-Compatible dataset, with varied budgets in the targeted setting.
Perturbation size is quantified in terms of $L_2$ and $L_{\infty}$ norms of the perturbations in RGB space ($\overline{L_2}$ and $\overline{L_{\infty}}$) and also in terms of image-level accumulated perceptual color difference ( $\overline{C_2}$).
For this relatively easy untargeted case, PerC-AL is initialized with $\alpha_c=0.1$.}
\label{tab:quan_1}
\end{table}
\end{document}